\pdfoutput=1

\documentclass[11pt]{article}

\usepackage{acl}

\usepackage{times}
\usepackage{latexsym}

\usepackage[T1]{fontenc}

\usepackage[utf8]{inputenc}

\usepackage{microtype}
\usepackage{url}            
\usepackage{booktabs}       
\usepackage{amsfonts}       
\usepackage{nicefrac}       
\usepackage{microtype}      
\usepackage{xcolor}         
\usepackage{amsmath}
\usepackage{multirow}
\usepackage{hyperref}
\usepackage{url}
\usepackage{subcaption}
\usepackage{graphicx}
\usepackage{float}
\usepackage{wrapfig,lipsum,booktabs}
\usepackage{wrapfig}
\restylefloat{figure}

\usepackage{amsthm}
\newtheorem{definition}{Definition}

\usepackage{booktabs,etoolbox}

\newlength{\toprulewidth}
\patchcmd{\toprule}
  {\heavyrulewidth}{\toprulewidth}
  {}{}

\patchcmd{\bottomrule}
  {\heavyrulewidth}{\toprulewidth}
  {}{}

\usepackage{algorithm} 
\usepackage{algorithmic}

\DeclareMathOperator*{\argmax}{argmax} 

\usepackage{listings}
\usepackage{xcolor}

\definecolor{codegreen}{rgb}{0,0.6,0}
\definecolor{codegray}{rgb}{0.5,0.5,0.5}
\definecolor{codepurple}{rgb}{0.58,0,0.82}
\definecolor{backcolour}{rgb}{0.95,0.95,0.92}

\lstdefinestyle{mystyle}{
    backgroundcolor=\color{backcolour},   
    commentstyle=\color{codegreen},
    keywordstyle=\color{magenta},
    numberstyle=\tiny\color{codegray},
    stringstyle=\color{codepurple},
    basicstyle=\ttfamily\tiny,
    breakatwhitespace=false,         
    breaklines=true,                 
    captionpos=b,                    
    keepspaces=true,                 
    showspaces=false,                
    showstringspaces=false,
    showtabs=false,                  
    tabsize=2
}

\title{Learning to Generalize to More: \\ Continuous Semantic Augmentation for Neural Machine Translation}

\author{Xiangpeng Wei\textsuperscript{\dag},
	Heng Yu\textsuperscript{\dag},
	Yue Hu\textsuperscript{\ddag,\S},
	Rongxiang Weng\textsuperscript{\dag},\\
	\textbf{Weihua Luo\textsuperscript{\dag},
	Jun Xie\textsuperscript{\dag},
	Rong Jin\textsuperscript{\dag}}\\
	\textsuperscript{\dag}Machine Intelligence Technology Lab, Alibaba DAMO Academy, Hangzhou, China\\
	\textsuperscript{\ddag}Institute of Information Engineering, Chinese Academy of Sciences, Beijing, China\\
	\textsuperscript{\S}School of Cyber Security, University of Chinese Academy of Sciences, Beijing, China\\
	\texttt{pemywei@gmail.com}\\
	\texttt{\url{https://github.com/pemywei/csanmt}}\\
}

\begin{document}
\maketitle
\begin{abstract}
The principal task in supervised neural machine translation (NMT) is to learn to generate target sentences conditioned on the source inputs from a set of parallel sentence pairs, and thus produce a model capable of generalizing to unseen instances. However, it is commonly observed that the generalization performance of the model is highly influenced by the amount of parallel data used in training. Although data augmentation is widely used to enrich the training data, conventional methods with discrete manipulations fail to generate diverse and faithful training 
samples. In this paper, we present a novel data augmentation paradigm termed Continuous Semantic Augmentation (\textsc{CsaNMT}), which augments each training instance with an adjacency semantic region that could cover adequate variants of literal expression under the same meaning. We conduct extensive experiments on both rich-resource and low-resource settings involving various language pairs, including WMT14 English$\rightarrow$\{German,French\}, NIST Chinese$\rightarrow$English and multiple low-resource IWSLT translation tasks. The provided empirical evidences show that \textsc{CsaNMT} sets a new level of performance among existing augmentation techniques, improving on the state-of-the-art by a large margin.\footnote{The core codes are contained in Appendix~\ref{appendix:codes}.}
\end{abstract}

\section{Introduction}
Neural machine translation (NMT) is one of the core topics in natural language processing, which aims to generate sequences of words in the target language conditioned on the source inputs~\citep{NIPS2014_5346,cho-etal-2014-learning,wu2016google,Vaswani2017Attention}. In the common supervised setting, the training objective is to learn a transformation from the source space to the target space $\mathcal{X} \mapsto \mathcal{Y}: f(\mathbf{y}|\mathbf{x};\Theta)$ with the usage of parallel data. In this way, NMT models are expected to be capable of generalizing to unseen instances with the help of large scale training data, which poses a big challenge for scenarios with limited resources.

To address this problem, various methods have been developed to leverage abundant unlabeled data for augmenting limited labeled data~\citep{sennrich-etal-2016-improving,cheng-etal-2016-semi,he2016dual,hoang-etal-2018-iterative,edunov-etal-2018-understanding,he2019revisiting,song2019mass}. For example, back-translation (BT)~\citep{sennrich-etal-2016-improving} makes use of the monolingual data on the target side to synthesize large scale pseudo parallel data, which is further combined with the real parallel corpus in machine translation task. Another line of research is to introduce adversarial inputs to improve the generalization of NMT models towards small perturbations~\citep{iyyer-etal-2015-deep,fadaee-etal-2017-data,wang-etal-2018-switchout,cheng-etal-2018-towards,gao-etal-2019-soft}. While these methods lead to significant boosts in translation quality, we argue that augmenting the observed training data in the discrete space inherently has two major limitations.

First, augmented training instances in discrete space are lack diversity. We still take BT as an example, it typically uses beam search~\citep{sennrich-etal-2016-improving} or greedy search~\citep{lample2017unsupervised,lample-etal-2018-phrase} to generate synthetic source sentences for each target monolingual sentence. Both of the above two search strategies identify the maximum a-posteriori (MAP) outputs~\citep{edunov-etal-2018-understanding}, and thus favor the most frequent one in case of ambiguity. \citet{edunov-etal-2018-understanding} proposed a sampling strategy from the output distribution to alleviate this issue, but this method typically yields synthesized data with low quality. While some extensions~\citep{wang-etal-2018-switchout,imamura-etal-2018-enhancement,khayrallah-etal-2020-simulated,NEURIPS2020_7221e5c8} augment each training instance with multiple literal forms, they still fail to cover adequate variants under the same meaning.

Second, it is difficult for augmented texts in discrete space to preserve their original meanings. In the context of natural language processing, discrete manipulations such as adds, drops, reorders, and/or replaces words in the original sentences often result in significant changes in semantics. To address this issue, \citet{gao-etal-2019-soft} and \citet{cheng-etal-2020-advaug} instead replace words with other words that are predicted using language model under the same context, by interpolating their embeddings. Although being effective, these techniques are limited to word-level manipulation and are unable to perform the whole sentence transformation, such as producing another sentence by rephrasing the original one so that they have the same meaning.

In this paper, we propose \textbf{C}ontinuous \textbf{S}emantic \textbf{A}ugmentation (\textsc{CsaNMT}), a novel data augmentation paradigm for NMT, to alleviate both limitations mentioned above. The principle of \textsc{CsaNMT} is to produce diverse training data from a semantically-preserved continuous space. Specifically, (1) we first train a semantic encoder via a \textit{tangential} contrast, which encourages each training instance to support an adjacency semantic region in continuous space and treats the tangent points of the region as the critical states of semantic equivalence. This is motivated by the intriguing observation made by recent work showing that the vectors in continuous space can easily cover adequate variants under the same meaning~\citep{wei-etal-2020-uncertainty}. (2) We then introduce a \textit{Mixed Gaussian Recurrent Chain} (\textsc{Mgrc}) algorithm to sample a cluster of vectors from the adjacency semantic region. 
(3) Each of the sampled vectors is finally incorporated into the decoder by developing a \textit{broadcasting integration network}, which is agnostic to model architectures. As a consequence, transforming discrete sentences into the continuous space can effectively augment the training data space and thus improve the generalization capability of NMT models.

We evaluate our framework on a variety of machine translation tasks, including WMT14 English-German/French, NIST Chinese-English and multiple IWSLT tasks. Specifically, \textsc{CsaNMT} sets the new state of the art among existing augmentation techniques on the WMT14 English-German task with \textbf{30.94} BLEU score. In addition, our approach could achieve comparable performance with the baseline model with the usage of only \textbf{25\%} of training data. This reveals that \textsc{CsaNMT} has great potential to achieve good results with very few data. Furthermore, \textsc{CsaNMT} demonstrates consistent improvements over strong baselines in low resource scenarios, such as IWSLT14 English-German and IWSLT17 English-French.

\section{Framework}\label{sec:method}


\paragraph{Problem Definition} Supposing $\mathcal{X}$ and $\mathcal{Y}$ are two data spaces that cover all possible sequences of words in source and target languages, respectively. We denote $(\mathbf{x},\mathbf{y}) \in (\mathcal{X},\mathcal{Y})$ as a pair of two sentences with the same meaning, where $\mathbf{x}=\{x_1,x_2,...,x_T\}$ is the source sentence with $T$ tokens, and $\mathbf{y}=\{y_1,y_2,...,y_{T^{\prime}}\}$ is the target sentence with $T^{\prime}$ tokens. A sequence-to-sequence model is usually applied to neural machine translation, which aims to learn a transformation from the source space to the target space $\mathcal{X} \mapsto \mathcal{Y}: f(\mathbf{y}|\mathbf{x};\Theta)$ with the usage of parallel data. Formally, given a set of observed sentence pairs $\mathcal{C}=\{(\mathbf{x}^{(n)},\mathbf{y}^{(n)})\}_{n=1}^{N}$, the training objective is to maximize the log-likelihood:
\begin{equation}
\footnotesize
\begin{split}
    J_{mle}(\Theta) &= \mathbb{E}_{(\mathbf{x},\mathbf{y}) \sim \mathcal{C}} \big( \log P(\mathbf{y}|\mathbf{x};\Theta) \big).
\end{split}
\label{eq:1}
\end{equation}
The log-probability is typically decomposed as: $\log P(\mathbf{y}|\mathbf{x};\Theta) = \sum_{t=1}^{T^{\prime}} \log P(y_t|\mathbf{y}_{<t},\mathbf{x};\Theta)$, where $\Theta$ is a set of trainable parameters and $\mathbf{y}_{<t}$ is a partial sequence before time-step $t$.

However, there is a major problem in the common supervised setting for neural machine translation, that is the number of training instances is very limited because of the cost in acquiring parallel data. This makes it difficult to learn an NMT model generalized well to unseen instances. Traditional data augmentation methods generate more training samples by applying discrete manipulations to unlabeled (or labeled) data, such as back-translation or randomly replacing a word with another one, which usually suffer from the problems of semantic deviation and the lack of diversity.

\begin{figure}
    \centering
    \includegraphics[scale=0.6]{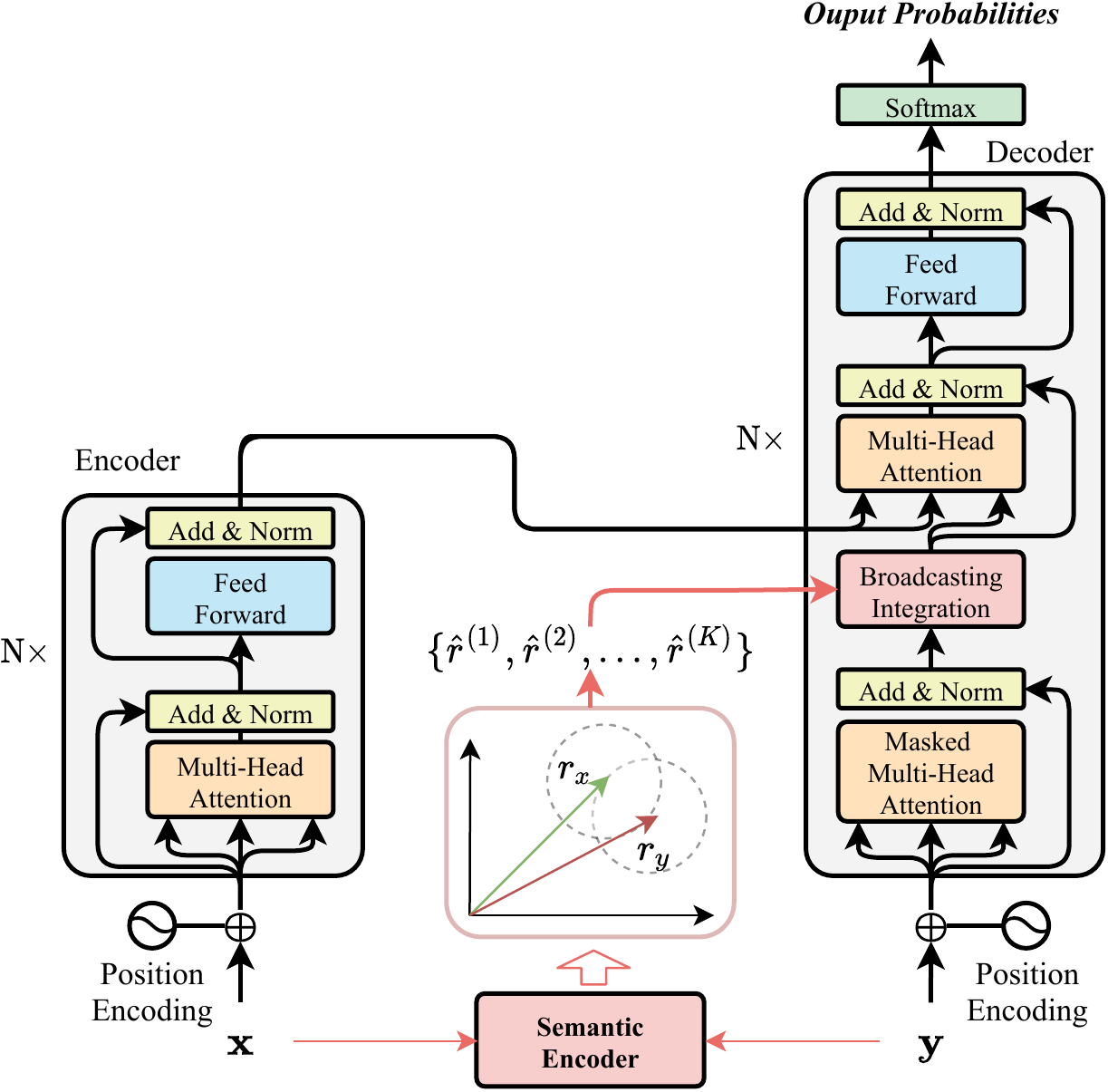}
    \caption{The framework of the \textsc{CsaNMT}.}
    \label{fig:model}
\end{figure}

\subsection{Continuous Semantic Augmentation}
We propose a novel data augmentation paradigm for neural machine translation, termed continuous semantic augmentation (\textsc{CsaNMT}), to better generalize the model's capability to unseen instances. We adopt the Transformer~\citep{Vaswani2017Attention} model as a backbone, and the framework is shown in Figure~\ref{fig:model}. In this architecture, an extra semantic encoder translates the source $\mathbf{x}$ and the target sentence $\mathbf{y}$ to real-value vectors $r_x=\psi(\mathbf{x};\Theta^{\prime})$ and $r_y=\psi(\mathbf{y};\Theta^{\prime})$ respectively, where $\psi(\cdot;\Theta^{\prime})$ is the forward function of the semantic encoder parameterized by $\Theta^{\prime}$ (parameters other than $\Theta$). 
\begin{definition}
  There is a universal semantic space among the source and the target languages for neural machine translation, which is established by a semantic encoder. It defines a forward function $\psi(\cdot;\Theta^{\prime})$ to map discrete sentences into continuous vectors, that satisfies: $\forall (\mathbf{x},\mathbf{y}) \in (\mathcal{X},\mathcal{Y}): r_x=r_y$. Besides, an adjacency semantic region $\nu(r_{x},r_{y})$ in the semantic space describes adequate variants of literal expression centered around each observed sentence pair $(\mathbf{x},\mathbf{y})$.
\end{definition}
In our scenario, we first sample a series of vectors (denoted by $\mathcal{R}$) from the adjacency semantic region to augment the current training instance, that is $\mathcal{R} = \{\hat{r}^{(1)},\hat{r}^{(2)},...,\hat{r}^{(K)}\}, \ {\rm where} \ \hat{r}^{(k)} \sim \nu(r_{x},r_{y})$. $K$ is the hyperparameter that determines the number of sampled vectors. Each sample $\hat{r}^{(k)}$ is then integrated into the generation process through a broadcasting integration network:
\begin{equation}
\footnotesize
    \hat{o}_t = W_1 \hat{r}^{(k)} + W_2 o_t + b,
\label{eq:2}
\end{equation}
where $o_t$ is the output of the self-attention module at position $t$. Finally, the training objective in Eq. (\ref{eq:1}) can be improved as
\begin{equation}
\footnotesize
\begin{split}
    J_{mle}(\Theta) =& \mathbb{E}_{(\mathbf{x},\mathbf{y}) \sim \mathcal{C},\hat{r}^{(k)} \in \mathcal{R}} \big(\log P(\mathbf{y}|\mathbf{x},\hat{r}^{(k)};\Theta)\big) \big).
\end{split}
\label{eq:mle}
\end{equation}
By augmenting the training instance $(\mathbf{x},\mathbf{y})$ with diverse samples from the adjacency semantic region, the model is expected to generalize to more unseen instances. To this end, we must consider such two problems: (1) \textbf{\textit{How to optimize the semantic encoder so that it produces a meaningful adjacency semantic region for each observed training pair.}} (2) \textbf{\textit{How to obtain samples from the adjacency semantic region in an efficient and effective way.}} In the rest part of this section, we introduce the resolutions of these two problems, respectively.

\begin{figure}
    \centering
    \includegraphics[scale=1.0]{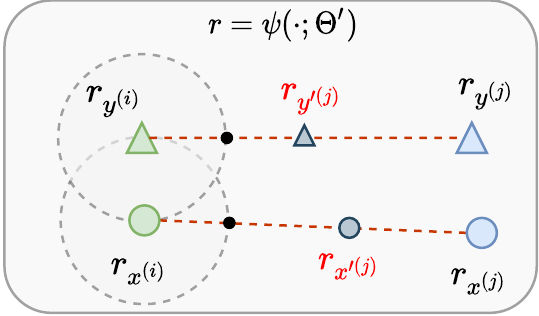}
    \caption{The diagram of formulating the adjacency semantic region for the sentence pair $(\mathbf{x}^{(i)},\mathbf{y}^{(i)})$.}
    \label{fig:ctl}
\end{figure}

\paragraph{Tangential Contrastive Learning} We start from analyzing the geometric interpretation of adjacency semantic regions. The schematic diagram is illustrated in Figure~\ref{fig:ctl}. Let $(\mathbf{x}^{(i)},\mathbf{y}^{(i)})$ and $(\mathbf{x}^{(j)},\mathbf{y}^{(j)})$ are two instances randomly sampled from the training corpora. For $(\mathbf{x}^{(i)},\mathbf{y}^{(i)})$, the adjacency semantic region $\nu(r_{x^{(i)}},r_{y^{(i)}})$ is defined as the union of two closed balls that are centered by $r_{x^{(i)}}$ and $r_{y^{(i)}}$, respectively. The radius of both balls is $d = \parallel r_{x^{(i)}} - r_{y^{(i)}} \parallel_{2}$, which is also considered as a slack variable for determining semantic equivalence. The underlying interpretation is that vectors whose distances from $r_{x^{(i)}}$ (or $r_{y^{(i)}}$) do not exceed $d$, are semantically-equivalent to both $r_{x^{(i)}}$ and $r_{y^{(i)}}$. To make $\nu(r_{x^{(i)}},r_{y^{(i)}})$ conform to the interpretation, we employ a similar method as in~\citep{Zheng_2019_CVPR,wei2021on} to optimize the semantic encoder with the \textit{tangential} contrast.

Specifically, we construct negative samples by applying the convex interpolation between the current instance and other ones in the same training batch for instance comparison. And the tangent points (i.e., the points on the boundary) are considered as the critical states of semantic equivalence. The training objective is formulated as:
\begin{equation}
\footnotesize
\begin{split}
J_{ctl}(\Theta^{\prime}) = &\mathbb{E}_{(\mathbf{x}^{(i)},\mathbf{y}^{(i)}) \sim \mathcal{B}} \bigg( \log \frac{e^{s \big( r_{x^{(i)}},r_{y^{(i)}} \big) }}{e^{s \big( r_{x^{(i)}},r_{y^{(i)}} \big) } + \xi} \bigg),\\
\xi = & \sum_{j \& j \ne i}^{|\mathcal{B}|} \Big ( e^{s \big( r_{y^{(i)}},r_{y^{\prime(j)}} \big) } + e^{s \big( r_{x^{(i)}},r_{x^{\prime(j)}} \big) } \Big),
\end{split}
\label{eq:ctl}
\end{equation}
where $\mathcal{B}$ indicates a batch of sentence pairs randomly selected from the training corpora $\mathcal{C}$, and $s(\cdot)$ is the score function that computes the cosine similarity between two vectors. The negative samples $r_{x^{\prime(j)}}$ and $r_{y^{\prime(j)}}$ are designed as the following interpolation:  
\begin{equation}
\footnotesize
    \begin{split}
        r_{x^{\prime(j)}} &= r_{x^{(i)}} + {\lambda}_x (r_{x^{(j)}}-r_{x^{(i)}}), {\lambda}_x \in (\frac{d}{d_x^{\prime}},1], \\
        r_{y^{\prime(j)}} &= r_{y^{(i)}} + {\lambda}_y (r_{y^{(j)}}-r_{y^{(i)}}), {\lambda}_y \in (\frac{d}{d_y^{\prime}},1],
    \end{split}
    \label{eq:interpolation}
\end{equation}
where $d_x^{\prime} = \parallel r_{x^{(i)}} - r_{x^{(j)}} \parallel_2$ and $d_y^{\prime} = \parallel r_{y^{(i)}} - r_{y^{(j)}} \parallel_2$. The two equations in Eq. (\ref{eq:interpolation}) set up when $d^{\prime}_x$ and $d^{\prime}_y$ are larger than $d$ respectively, or else $r_{x^{\prime(j)}} = r_{x^{(j)}}$ and $r_{y^{\prime(j)}} = r_{y^{(j)}}$. 
According to this design, an adjacency semantic region for the $i$-th training instance can be fully established by interpolating various instances in the same training batch. We follow~\citet{wei2021on} to adaptively adjust the value of $\lambda_x$ (or $\lambda_y$) during the training process, and refer to the original paper for details.

\begin{figure}
    \centering
    \includegraphics[scale=1.4]{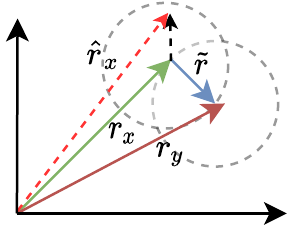}
    \caption{The geometric diagram of the proposed \textsc{Mgrc} sampling. $r_x$ and $r_y$ are the representations of the source sentence $\mathbf{x}$ and the target sentence $\mathbf{y}$, respectively. To construct the augmented sample, a straightforward idea is that: (1) transform the norm or the direction of $\tilde{r}=r_y - r_x$, formulated as $\omega \odot \tilde{r}$ (e.g., the black dashed arrow), in which each element $\omega_i \in [-1,1]$, and (2) combine $r_x$ (or $r_y$) and the transformation $\omega \odot \tilde{r}$ as $\hat{r}_x = r_x + \omega \odot \tilde{r}$ (i.e., the red dashed arrow).}
    \label{fig:sampling}
\end{figure}


\begin{table*}[t]
\begin{center}
\resizebox{\textwidth}{!}{
\begin{tabular}{lrccccccc}
\toprule
\textbf{Method} & \textbf{\#Params.} & \textbf{Valid.} & \textbf{MT02} & \textbf{MT03} & \textbf{MT04} & \textbf{MT05} & \textbf{MT08} & \textbf{Avg.}\\
\midrule
Transformer, base (\textit{our implementation}) & 84M & 45.09 & 45.63 & 45.07 & 46.59 & 45.84 & 36.18 & 43.86\\
Back-translation~\citep{sennrich-etal-2016-improving}$^*$ & 84M & 46.71 & 47.22 & 46.86 & 47.36 & 46.65 & 36.69 & 44.96\\
SwitchOut~\citep{wang-etal-2018-switchout}$^*$ & 84M & 46.13 & 46.72 & 45.69 & 47.08 & 46.19 & 36.47 & 44.43\\
SemAug~\citep{wei-etal-2020-uncertainty} & 86M & - & - & - & 49.15 & 49.21 & 40.94 & -\\
AdvAug~\citep{cheng-etal-2020-advaug} & - & 49.26 & 49.03 & 47.96 & 48.86 & 49.88 & 39.63 & 47.07\\
\midrule
\textsc{CsaNMT}, base & 96M & \bf 50.46 & \bf 49.65 & \bf 48.84 & \bf 49.80 & \bf 50.40 & \bf 41.63 & \bf 48.06\\
\bottomrule
\end{tabular}}
\end{center}
\setlength{\abovecaptionskip}{0.1cm}
\caption{BLEU scores [\%] on Zh$\rightarrow$En translation. ``\textbf{Params.}'' denotes the number of parameters (M=million). ``$*$'' indicates the results obtained by our implementation, we construct multiple pseudo sources for each target during back-translation but rather introducing extra monolingual corpora as in~\citep{wei-etal-2020-uncertainty} for fairer comparisons.}
\label{nist-table}
\end{table*}

\paragraph{\textsc{Mgrc} Sampling} To obtain augmented data from the adjacency semantic region for the training instance $(\mathbf{x},\mathbf{y})$, we introduce a Mixed Gaussian Recurrent Chain (denoted by \textsc{Mgrc}) algorithm to design an efficient and effective sampling strategy. As illustrated in Figure~\ref{fig:sampling}, we first transform the bias vector $\tilde{r} = r_y - r_x$ according to a pre-defined scale vector $\omega$, that is $\omega \odot \tilde{r}$, where $\odot$ is the element-wise product operation. Then, we construct a novel sample $\hat{r} = r + \omega \odot \tilde{r}$ for augmenting the current instance, in which $r$ is either $r_x$ or $r_y$. As a consequence, the goal of the sampling strategy turns into find a set of scale vectors, i.e. $\omega \in \{\omega^{(1)},\omega^{(2)},...,\omega^{(K)}\}$. Intuitively, we can assume that $\omega$ follows a distribution with universal or Gaussian forms, despite the latter demonstrates better results in our experience. Formally, we design a mixed Gaussian distribution as follow:
\begin{equation}
\footnotesize
\begin{split}
    \omega^{(k)} &\sim p(\omega|\omega^{(1)},\omega^{(2)},...,\omega^{(k-1)}),\\
    p &= \eta \mathcal{N}\big(\mathbf{0},{\rm diag}(\mathcal{W}_{r}^2) \big)\\
    &+ (1.0-\eta) \mathcal{N}\bigg(\frac{1}{k-1}\sum_{i=1}^{k-1}\omega^{(i)},\mathbf{1} \bigg).
\end{split}
\label{eq:mcmc}
\end{equation}
This framework unifies the recurrent chain and the rejection sampling mechanism. Concretely, we first normalize the importance of each dimension in $\tilde{r}$ as ${\mathcal{W}_{r}=\frac{|\tilde{r}|-min(|\tilde{r}|)}{max(|\tilde{r}|)-min(|\tilde{r}|)}}$, the operation $|\cdot|$ takes the absolute value of each element in the vector, which means the larger the value of an element is the more informative it is. Thus $\mathcal{N}(\mathbf{0},{\rm diag}(\mathcal{W}_{r}^2))$ limits the range of sampling to a subspace of the adjacency semantic region, and rejects to conduct sampling from the uninformative dimensions. Moreover, $\mathcal{N}(\frac{1}{k-1}\sum_{i=1}^{k-1}\omega^{(i)},\mathbf{1})$ simulates a recurrent chain that generates a sequence of reasonable vectors where the current one is dependent on the prior vectors. The reason for this design is that we expect that $p$ in Eq. (\ref{eq:mcmc}) can become a stationary distribution with the increase of the number of samples, which describes the fact that the diversity of each training instance is not infinite. $\eta$ is a hyper-parameter to balance the importance of the above two Gaussian forms. For a clearer presentation, Algorithm~\ref{alg:1} summarizes the sampling process.

\begin{algorithm}[t]
\footnotesize
	\caption{\textsc{Mgrc} Sampling}
	\label{alg:1}
	\begin{algorithmic}[1]
	    \REQUIRE The representations of the training instance $(\mathbf{x},\mathbf{y})$, i.e. $r_x$ and $r_y$.
	    \ENSURE A set of augmented samples $\mathcal{R} = \{\hat{r}^{(1)},\hat{r}^{(2)},...,\hat{r}^{(K)}\}$
		\STATE Normalizing the importance of each element in $\tilde{r}=r_y-r_x$: ${\mathcal{W}_{r}=\frac{|\tilde{r}|-min(|\tilde{r}|)}{max(|\tilde{r}|)-min(|\tilde{r}|)}}$
		\STATE Set $k=1$, $\omega^{(1)} \sim \mathcal{N}(\mathbf{0},{\rm diag}(\mathcal{W}_{r}^2))$, $\hat{r}^{(1)} = r + \omega^{(1)} \odot (r_y - r_x)$
		\STATE Initialize the set of samples as $\mathcal{R}=\{\hat{r}^{(1)}\}$.
	    \WHILE{$k \le (K-1)$}
	            \STATE $k \leftarrow k+1$
    	        \STATE Calculate the current scale vector: $\omega^{(k)} \sim p(\omega|\omega^{(1)},\omega^{(2)},...,\omega^{(k-1)}$ according to Eq. (\ref{eq:mcmc}).
    	        \STATE Calculate the current sample: $\hat{r}^{(k)} = r + \omega^{(k)} \odot (r_y - r_x)$.
	            \STATE $\mathcal{R} \leftarrow \mathcal{R} \bigcup \{\hat{r}^{(k)}\}$.
	    \ENDWHILE
	\end{algorithmic}
\end{algorithm}

\subsection{Training and Inference}
The training objective in our approach is a combination of $J_{mle}(\Theta)$ in Eq. (\ref{eq:mle}) and $J_{ctl}(\Theta^{\prime})$ in Eq. (\ref{eq:ctl}). In practice, we introduce a two-phase training procedure with mini-batch losses. Firstly, we train the semantic encoder from scratch using the task-specific data, i.e. $\Theta^{\prime *} = \argmax_{\Theta^{\prime}} J_{ctl}(\Theta^{\prime})$. 
Secondly, we optimize the encoder-decoder model by maximizing the log-likelihood, i.e. $\Theta^{*} = \argmax_{\Theta} J_{mle}(\Theta)$, and fine-tune the semantic encoder with a small learning rate at the same time.

During inference, the sequence of target words is generated auto-regressively, which is almost the same as the vanilla Transformer~\citep{Vaswani2017Attention}. A major difference is that our method involves the semantic vector of the input sequence for generation: $y_t^* = \argmax_{y_t}P(\cdot|\mathbf{y}_{<t},\mathbf{x},r_x;\Theta)$, where $r_x = \psi(\mathbf{x};\Theta^{\prime})$. This module is plug-in-use as well as is agnostic to model architectures.


\begin{table*}[t]
  \begin{center}
  \resizebox{\textwidth}{!}{
  \begin{tabular}{lrccrcc}
  \toprule
  \multirow{2}{*}{\textbf{Model}}& \multicolumn{3}{c }{\textbf{WMT 2014 En$\rightarrow$De}}  & \multicolumn{3}{c}{\textbf{WMT 2014 En$\rightarrow$Fr}} \\
  \cmidrule(r){2-4} \cmidrule(r){5-7}
  & \bf \#Params. & \bf BLEU & \bf SacreBLEU & \bf \#Params. & \bf BLEU & \bf SacreBLEU\\
    \midrule
    Transformer, base (\textit{our implementation}) & 62M & 27.67 & 26.8 & 67M & 40.53 & 38.5 \\
     Transformer, big (\textit{our implementation}) & 213M & 28.79 & 27.7 & 222M & 42.36 & 40.3 \\
    Back-Translation~\citep{sennrich-etal-2016-improving}$^*$ & 213M & 29.25 & 28.2 & 222M & 41.73 & 39.7 \\
    SwitchOut~\citep{wang-etal-2018-switchout}$^*$ & 213M & 29.18 & 28.1 & 222M & 41.62 & 39.6 \\
    SemAug~\citep{wei-etal-2020-uncertainty} & 221M & 30.29 & - & 230M & 42.92 & - \\
    AdvAug~\citep{cheng-etal-2020-advaug} & $^\dag$65M & 29.57 & - & - & - & - \\
    Data Diversification~\citep{NEURIPS2020_7221e5c8} & $^\dag$1260M & 30.70 & - & $^\dag$1332M & \bf 43.70 & - \\
    \midrule
    \textsc{CsaNMT}, base & 74M & 30.16 & 29.2 & 80M & 42.40 & 40.3 \\
    \textsc{CsaNMT}, big & 265M & \bf 30.94 & \bf 29.8 & 274M & \bf 43.68 & \bf 41.6 \\
  \bottomrule
\end{tabular}}
\end{center}
\setlength{\abovecaptionskip}{0.25cm}
\caption{BLEU scores [\%] on the WMT14 En$\rightarrow$De and En$\rightarrow$Fr tasks. ``$*$'' indicates the results obtained by our implementation, which is the same in Table~\ref{nist-table}. ``$^\dag$'' denote estimate values. We further compare against the baselines with increased amounts of parameters, and investigate the performance of \textsc{CsaNMT} equipped with much stronger baselines (e.g. deep and scale Transformers~\citep{ott-etal-2018-scaling,wang-etal-2019-learning-deep,wei-etal-2020-multiscale}) in \textbf{Sec.~\ref{sec:analysis}}.}
\label{wmt-table}
\end{table*}

\section{Experiments}

We first apply \textsc{CsaNMT} to NIST Chinese-English (Zh$\rightarrow$En), WMT14 English-German (En$\rightarrow$De) and English-French (En$\rightarrow$Fr) tasks, and conduct extensive analyses for better understanding the proposed method. And then we generalize the capability of our method to low-resource IWSLT tasks.

\subsection{Settings}

\textbf{Datasets.} For the Zh$\rightarrow$En task, the LDC corpus is taken into consideration, which consists of 1.25M sentence pairs with 27.9M Chinese words and 34.5M English words, respectively. The NIST 2006 dataset is used as the validation set for selecting the best model, and NIST 2002 (MT02), 2003 (MT03), 2004 (MT04), 2005 (MT05), 2008 (MT08) are used as the test sets. For the En$\rightarrow$De task, we employ the popular WMT14 dataset, which consists of approximately 4.5M sentence pairs for training. We select \texttt{newstest2013} as the validation set and \texttt{newstest2014} as the test set. For the En$\rightarrow$Fr task, we use the significantly larger WMT14 dataset consisting of 36M sentence pairs. The combination of \{\texttt{newstest2012, 2013}\} was used for model selection and the experimental results were reported on \texttt{newstest2014}. Refer to \textbf{Appendix~\ref{appendix:rich-dataset}} for more details.

\textbf{Training Details.} We implement our approach on top of the Transformer~\citep{Vaswani2017Attention}. 
The semantic encoder is a 4-layer transformer encoder with the same hidden size as the backbone model. Following sentence-bert~\citep{reimers-gurevych-2019-sentence}, we average the outputs of all positions as the sequence-level representation. The learning rate for finetuning the semantic encoder at the second training stage is set as $1e-5$. All experiments are performed on 8 V100 GPUs. We accumulate the gradient of 8 iterations and update the models with a batch of about 65K tokens. The hyperparameters $K$ and $\eta$ in \textsc{Mgrc} sampling are tuned on the validation set with the range of $K \in \{10,20,40,80\}$ and $\eta \in \{0.15,0.30,0.45,0.6,0.75,0.90\}$. We use the default setup of $K=40$ for all three tasks, $\eta = 0.6$ for both Zh$\rightarrow$En and En$\rightarrow$De while $\eta = 0.45$ for En$\rightarrow$Fr. For evaluation, the beam size and length penalty are set to 4 and 0.6 for the En$\rightarrow$De as well as En$\rightarrow$Fr, while 5 and 1.0 for the Zh$\rightarrow$En task.


\begin{figure*}
    \centering
    \begin{subfigure}[b]{0.22\textwidth}
        \includegraphics[scale=0.48]{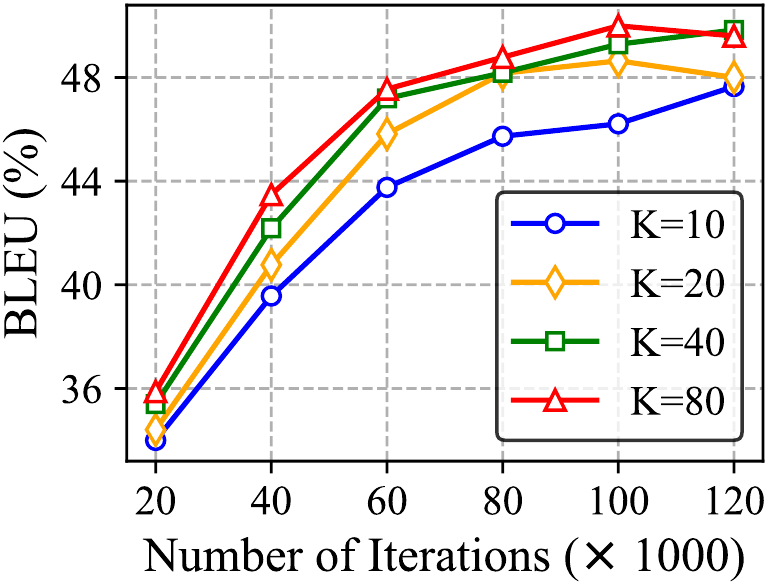}
        \caption{NIST Zh$\rightarrow$En}
        \label{fig:k-zhen}
    \end{subfigure}
    \hspace{0.32cm}
    \begin{subfigure}[b]{0.22\textwidth}
        \includegraphics[scale=0.39]{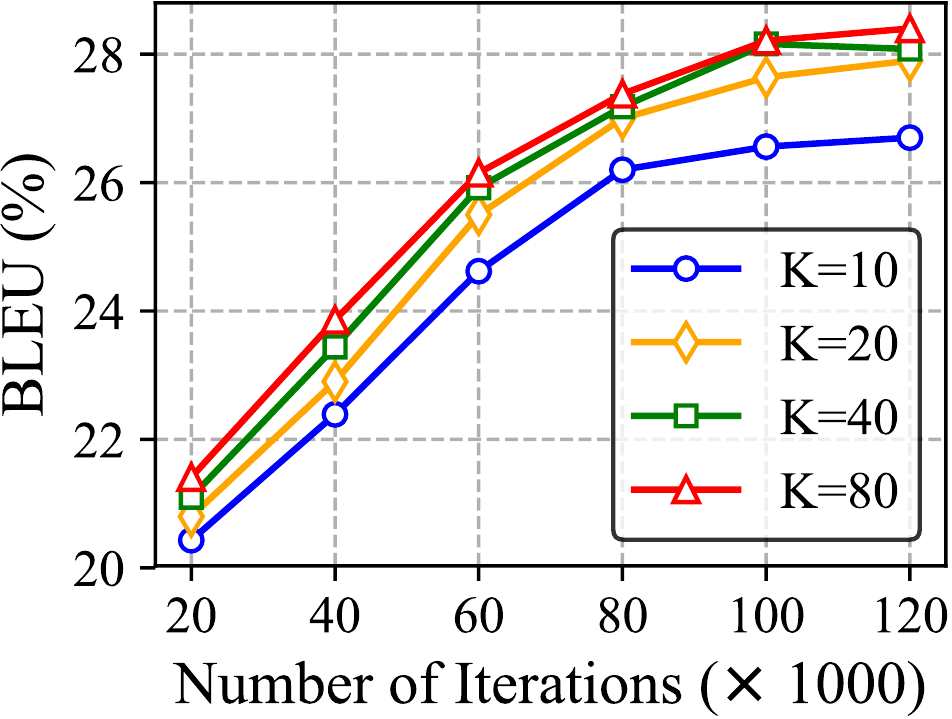}
        \caption{WMT14 En$\rightarrow$De}
        \label{fig:k-ende}
    \end{subfigure}
    \hspace{0.32cm}
    \begin{subfigure}[b]{0.22\textwidth}
        \includegraphics[scale=0.485]{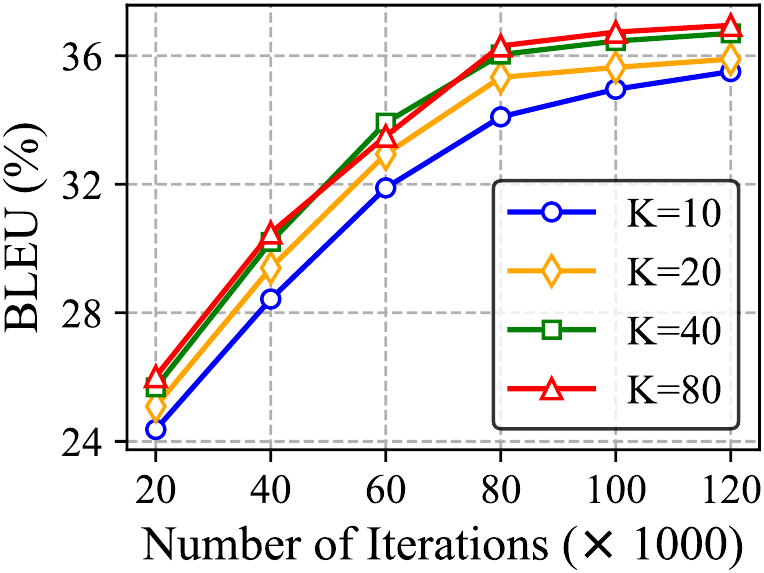}
        \caption{WMT14 En$\rightarrow$Fr}
        \label{fig:k-enfr}
    \end{subfigure}
    \hspace{0.32cm}
    \begin{subfigure}[b]{0.22\textwidth}
        \includegraphics[scale=0.38]{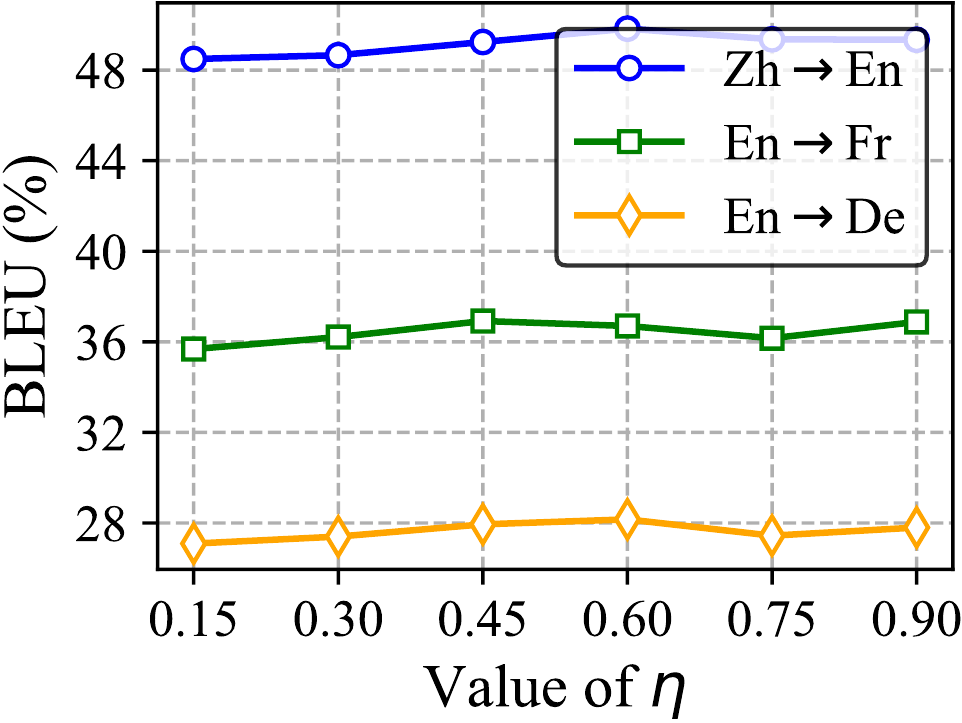}
        \caption{Effect of $\eta$.}
        \label{fig:eta}
    \end{subfigure}
    \caption{Effects of $K$ and $\eta$ on validation sets. (a), (b) and (c) depict the BLEU curves with different values of $K$ on Zh$\rightarrow$En, En$\rightarrow$De and En$\rightarrow$Fr, respectively. (d) demonstrates the performances of $\eta$ with different values.}
    \label{fig:k-eta}
\end{figure*}

\subsection{Main Results}

\textbf{Results of Zh$\rightarrow$En.} Table~\ref{nist-table} shows the results on the Chinese-to-English translation task. From the results, we can conclude that our approach outperforms existing augmentation strategies such as back-translation~\citep{sennrich-etal-2016-improving,wei-etal-2020-uncertainty} and switchout~\citep{wang-etal-2018-switchout} by a large margin (up to \textbf{3.63} BLEU), which verifies that augmentation in continuous space is more effective than methods with discrete manipulations. Compared to the approaches that replace words in the embedding space~\citep{cheng-etal-2020-advaug}, our approach also demonstrates superior performance, which reveals that sentence-level augmentation with continuous semantics works better on generalizing to unseen instances. Moreover, compared to the vanilla Transformer, our approach consistently achieves promising improvements on five test sets. 

\textbf{Results of En$\rightarrow$De and En$\rightarrow$Fr.} From Table~\ref{wmt-table}, our approach consistently performs better than existing methods~\citep{sennrich-etal-2016-improving,wang-etal-2018-switchout, wei-etal-2020-uncertainty,cheng-etal-2020-advaug}, yielding significant gains (0.65$\sim$1.76 BLEU) on the En$\rightarrow$De and En$\rightarrow$Fr tasks. An exception is that \citet{NEURIPS2020_7221e5c8} achieved comparable results with ours via multiple forward and backward NMT models, thus data diversification intuitively demonstrates lower training efficiency. Moreover, we observe that \textsc{CsaNMT} gives 30.16 BLEU on the En$\rightarrow$De task with the \texttt{base} setting, significantly outperforming the vanilla Transformer by 2.49 BLEU points. Our approach yields a further improvement of 0.68 BLEU by equipped with the wider architecture, demonstrating superiority over the standard Transformer by 2.15 BLEU. Similar observations can be drawn for the En$\rightarrow$Fr task.

\subsection{Analysis}
\label{sec:analysis}

\textbf{Effects of $K$ and $\eta$.} Figure~\ref{fig:k-eta} illustrates how the hyper-parameters $K$ and $\eta$ in \textsc{Mgrc} sampling affect the translation quality. From Figures 4(a)-4(c), we can observe that gradually increasing the number of samples significantly improves BLEU scores, which demonstrates large gaps between $K=10$ and $K=40$. However, assigning larger values (e.g., $80$) to $K$ does not result in further improvements among all three tasks. We conjecture that the reasons are two folds: (1) it is fact that the diversity of each training instance is not infinite and thus \textsc{Mgrc} gets saturated is inevitable with $K$ increasing. (2) \textsc{Mgrc} sampling with a scaled item (i.e., $\mathcal{W}_{r}$) may degenerate to traverse in the same place. This prompts us to design more sophisticated algorithms in future work. In our experiments, we default set $K=40$ to achieve a balance between the training efficiency and translation quality. Figure 4(d) shows the effect of $\eta$ on validation sets, which balances the importance of two Gaussian forms during the sampling process. The setting of $\eta=0.6$ achieves the best results on both the Zh$\rightarrow$En and En$\rightarrow$De tasks, and $\eta=0.45$ consistently outperforms other values on the En$\rightarrow$Fr task.

\textbf{Lexical diversity and semantic faithfulness.} 
We demonstrate both the lexical diversity (measured by TTR$=\frac{num. \ of \ types}{num. \ of \ tokens}$) of various translations and the semantic faithfulness of machine translated ones (measured by BLEURT with considering human translations as the references) in Table~\ref{tbl:ttr-bleurt}. It is clear that \textsc{CsaNMT} substantially bridge the gap of the lexical diversity between translations produced by human and machine. Meanwhile, \textsc{CsaNMT} shows a better capability on preserving the semantics of the generated translations than Transformer. We intuitively attribute the significantly increases of BLEU scores on all datasets to these two factors. We also have studied the robustness of \textsc{CsaNMT} towards noisy inputs and the translationese effect, see \textbf{Appendix~\ref{appendix:robustness}} for details.


\begin{figure}
    \centering
    \includegraphics[scale=0.70]{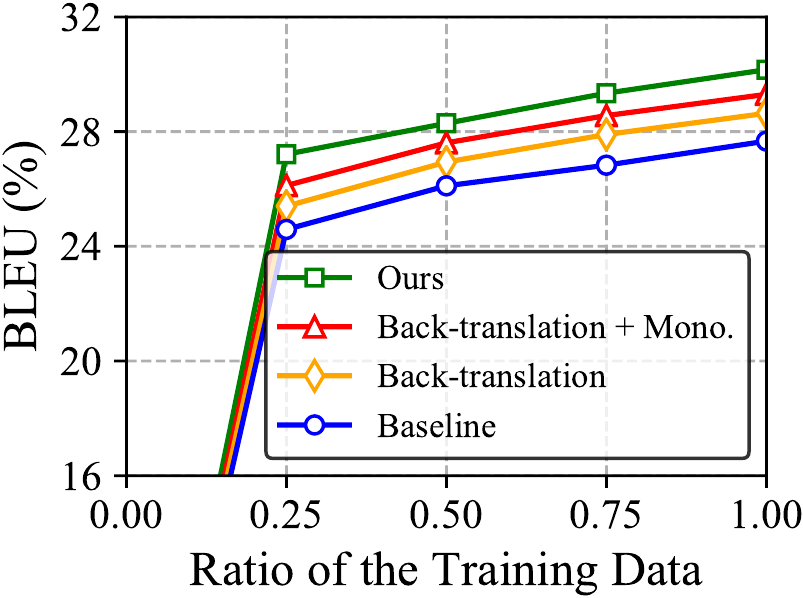}
\setlength{\abovecaptionskip}{0.1cm}
\caption{Comparison between discrete and continuous augmentations with different ratios of the training data.}
\label{fig:5}
\end{figure}

\begin{table}[t]
  \begin{center}
  \resizebox{0.48\textwidth}{!}{
  \begin{tabular}{lcc}
  \toprule
  \textbf{Model} & \textbf{BLEU} & \textbf{Dec. speed}\\
    \midrule
    Transformer-base & 27.67 & reference \\
    Default 4-layer semantic encoder & 30.16 & 0.95$\times$\\
    \midrule
    Remove the extra semantic encoder & 28.71 & 1.0$\times$\\
    Take PTMs as the semantic encoder & 31.10 & 0.62$\times$\\
  \bottomrule
\end{tabular}}
\end{center}
\setlength{\abovecaptionskip}{0.1cm}
\caption{Effect of the semantic encoder variants.}
\label{tbl:semantic-encoder}
\end{table}

\begin{table}[t]
  \begin{center}
  \resizebox{0.48\textwidth}{!}{
  \begin{tabular}{l|ccc|ccc}
  \toprule
    \multirow{2}{*}{} & \multicolumn{3}{c|}{\bf TTR} &
    \multicolumn{3}{c}{\bf BLEURT Score}\\
& Zh & De & Fr & Zh & De & Fr\\
    \midrule
    Human & 7.58\% & 22.08\% & 13.98\% & - & - & -\\
    \midrule
    Trans. & 6.95\% & 20.32\% & 11.76\% & 0.570 & 0.635 & 0.696\\
    \textsc{CsaNMT} & \bf 7.13\% & \bf 21.26\% & \bf 12.91\% & \bf 0.581 & \bf 0.684 & \bf 0.739\\
  \bottomrule
\end{tabular}}
\end{center}
\setlength{\abovecaptionskip}{0.1cm}
\caption{TTR (Type-Token-Ratio)~\citep{templin1957certain} and BLEURT scores of Zh$\rightarrow$En and En$\rightarrow$X translations produced by Human, vanilla Transformer (written as Trans.), and \textsc{CsaNMT}. ``Human'' translations mean the \textit{references} contained in the standard test sets. Refer to \textbf{Appendix~\ref{appendix:robustness}} for the results on robustness test sets.}
\label{tbl:ttr-bleurt}
\end{table}

\begin{table}[t]
  \begin{center}
  \resizebox{0.48\textwidth}{!}{
  \begin{tabular}{llcc}
  \toprule
  \# & \textbf{Objective} & \textbf{Sampling} & \textbf{BLEU}\\
    \midrule
    1 & Default tangential CTL & Default \textsc{Mgrc} & 30.16\\
    2 & Default tangential CTL & \textsc{Mgrc} w/o recurrent chain & 29.64\\
    3 & Default tangential CTL & \textsc{Mgrc} w/ uniform dist. & 29.78\\
    \midrule
    4 & Variational Inference & Gaussian sampling & 28.07\\
    5 & Cosine similarity & Default \textsc{Mgrc} & 28.19\\
  \bottomrule
\end{tabular}}
\end{center}
\setlength{\abovecaptionskip}{0.1cm}
\caption{Effect of \textsc{Mgrc} sampling and tangential contrastive learning on En$\rightarrow$De validation set.}
\label{tbl:imcmc-ctl}
\end{table}

\textbf{Effect of the semantic encoder.} We introduce two variants of the semantic encoder to investigate its performance on En$\rightarrow$De validation set. Specifically, (1) we remove the extra semantic encoder and construct the sentence-level representations by averaging the sequence of outputs of the vanilla sentence encoder. (2) We replace the default 4-layer semantic encoder with a large pre-trained model (PTM) (i.e., \textsc{XLM-R}~\cite{conneau-etal-2020-unsupervised}). The results are reported in Table~\ref{tbl:semantic-encoder}. Comparing line 2 with line 3, we can conclude that an extra semantic encoder is necessary for constructing the universal continuous space among different languages. Moreover, when the large PTM is incorporated, our approach yields further improvements, but it causes massive computational overhead. 

\textbf{Comparison between discrete and continuous augmentations.} To conduct detailed comparisons between different augmentation methods, we asymptotically increase the training data to analyze the performance of them on the En$\rightarrow$De translation. As in Figure~\ref{fig:5}, our approach significantly outperforms the back-translation method on each subset, whether or not extra monolingual data~\cite{sennrich-etal-2016-improving} is introduced. These results demonstrate the stronger ability of our approach than discrete augmentation methods on generalizing to unseen instances with the same set of observed data points. Encouragingly, our approach achieves comparable performance with the baseline model with only 25\% of training data, which  indicates that our approach has great potential to achieve good results with very few data.

\textbf{Effect of \textsc{Mgrc} sampling and tangential contrastive learning.} To better understand the effectiveness of the \textsc{Mgrc} sampling and the tangential contrastive learning, we conduct detailed ablation studies in Table~\ref{tbl:imcmc-ctl}. The details of four variants with different objectives or sampling strategies are shown in \textbf{Appendix~\ref{appendix:objective-sampling}}. From the results, we can observe that both removing the recurrent dependence and replacing the Gaussian forms with uniform distributions make the translation quality decline, but the former demonstrates more drops. We also have tried the training objectives with other forms, such as variational inference and cosine similarity, to optimize the semantic encoder. However, the BLEU score drops significantly.

\begin{figure}
    \centering
    \includegraphics[scale=0.4]{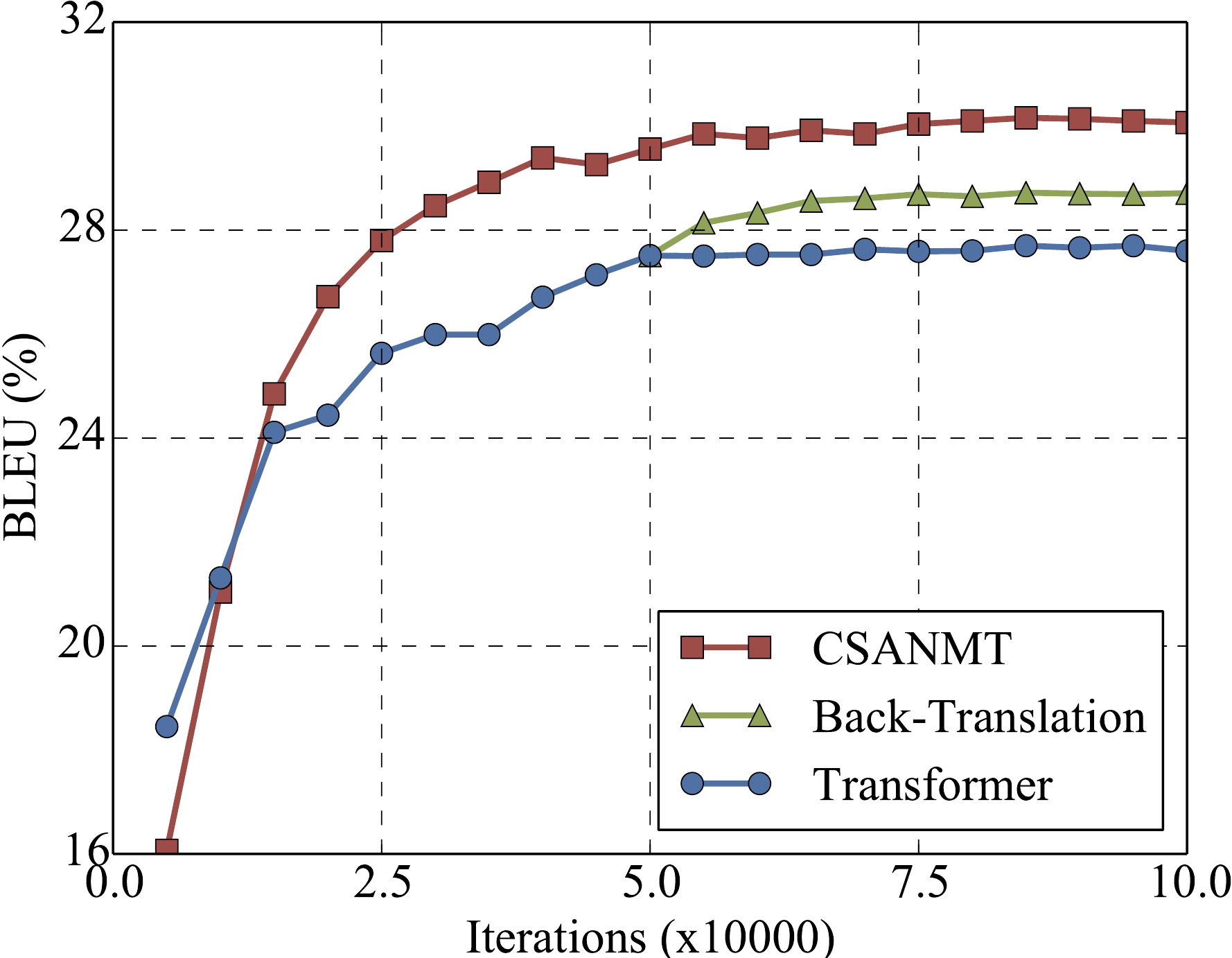}
\caption{BLEU curves over iterations on the WMT14 English$\rightarrow$German test set. Note that back-translation is initialized from the vanilla Transformer.}
\label{fig:bleu-curve}
\end{figure}

\textbf{Training Cost and Convergence.} Figure~\ref{fig:bleu-curve} shows the evolution of BLEU scores during training. It is obvious that our method performs consistently better than both the vanilla Transformer and the back-translation method at each iteration (except for the first 10K warm-up iterations, where the former one has access to less unique training data than the latter two due to the $K$ times over-sampling). For the vanilla Transformer, the BLEU score reaches its peak at about 52K iterations. In comparison, both \textsc{CsaNMT} and the back-translation method require 75K updates for convergence. In other words, \textsc{CsaNMT} spends 44\% more training costs than the vanilla Transformer, due to the longer time to make the NMT model converge with augmented training instances. This is the same as the back-translation method.

\begin{figure}
    \centering
    \includegraphics[scale=0.44]{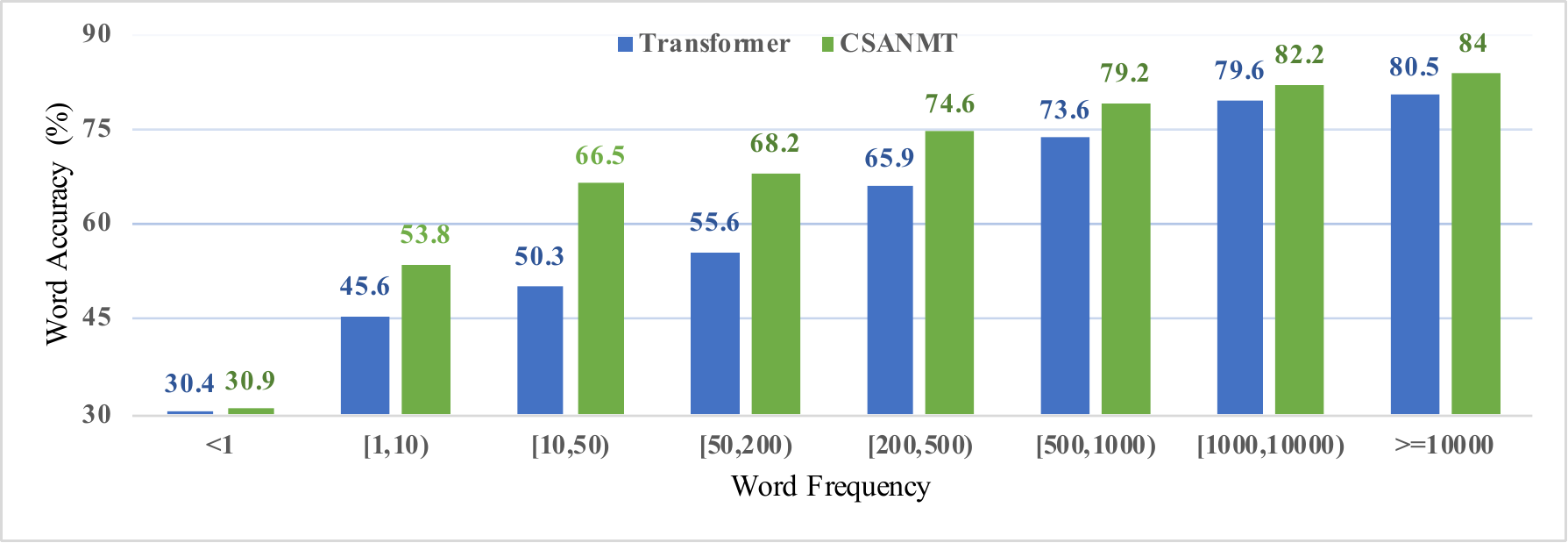}
\setlength{\abovecaptionskip}{0.1cm}
\caption{Comparison between the vanilla Transformer and \textsc{CsaNMT} on prediction accuracy of words with different frequencies.}
\label{fig:freq-acc}
\end{figure}

\textbf{Word prediction accuracy.} Figure~\ref{fig:freq-acc} illustrates the prediction accuracy of both frequent and rare words. As expected, \textsc{CsaNMT} generalizes to rare words better than the vanilla Transformer, and the gap of word prediction accuracy is as large as 16\%. This indicates that the NMT model alleviates the probability under-estimation of rare words via continuous semantic augmentation. 

\textbf{Effects of Additional Parameters and Strong Baselines.} In contrast to the vanilla Transformer, \textsc{CsaNMT} involves with approximate 20\% additional parameters. In this section, we further compare against the baselines with increased amounts of parameters, and investigate the performance of \textsc{CsaNMT} equipped with much stronger baselines (e.g. deep and scale Transformers~\citep{ott-etal-2018-scaling,wang-etal-2019-learning-deep,wei-etal-2020-multiscale}). From the results on WMT14 testsets in Table~\ref{tbl:strong-baselines}, we can observe that \textsc{CsaNMT} still outperforms the vanilla Transformer (by more than 1.2 BLEU) under the same amount of parameters, which shows that the additional parameters are not the key to the improvement. Moreover, \textsc{CsaNMT} yields at least 0.9 BLEU gains equipped with much stronger baselines. For example, the scale Transformer~\citep{ott-etal-2018-scaling}, which originally gives 29.3 BLEU in the En$\rightarrow$De task, now gives 31.37 BLEU with our continuous semantic augmentation strategy. It is important to mention that our method can help models to achieve further improvement, even if they are strong enough.

\begin{table}[t]
  \begin{center}
  \resizebox{0.48\textwidth}{!}{
  \begin{tabular}{lrcc}
  \toprule
  \textbf{Model} & \textbf{\#Params.} & \textbf{En$\rightarrow$De} & \textbf{En$\rightarrow$Fr}\\
    \midrule
    Transformer~\citep{Vaswani2017Attention}$^\dag$ & 213M & 28.40 & 41.80\\
    Transformer (\textit{our impl.}) & 213M & 28.79 & 42.36\\
    Transformer (\textit{our impl., 10 layers}) & 265M & 29.08 & 42.49\\
    \textsc{CsaNMT} & 265M & \bf 30.94 & \bf 43.68\\
    \midrule
    Scale Trans.~\citep{ott-etal-2018-scaling}$^\dag$ & 210M & 29.30 & 43.20\\
    \textsc{Deep}~\citep{wang-etal-2019-learning-deep} & 350M & 30.26 & 43.24\\
    \textsc{Msc}~\citep{wei-etal-2020-multiscale}$^\dag$ & 512M & 30.56 & - \\
    \midrule
    \textbf{Our \textsc{CsaNMT} with} & & & \\
    Scale Trans.~\citep{ott-etal-2018-scaling} & 263M & 31.37 & \bf 44.12\\
    \textsc{Deep}~\citep{wang-etal-2019-learning-deep} & 405M & 31.35 & -\\
    \textsc{Msc}~\citep{wei-etal-2020-multiscale} & 566M & \bf 31.49 & - \\
  \bottomrule
\end{tabular}}
\end{center}
\caption{BLEU scores [\%] on WMT14 testsets for the English-German (En$\rightarrow$De) and English-French (En$\rightarrow$Fr) tasks. Superscript $^\dag$ denotes the numbers are reported from the paper, others are based on our runs. ``-'' means omitted results because of the limitations of GPU resources. ``10 layers'' means that we construct the Transformer with a 10-layer encoder, thus it has the same amount of parameters as our model.}
\label{tbl:strong-baselines}
\end{table}

\subsection{Low-Resource Machine Translation}

\begin{table}[t]
  \begin{center}
  \resizebox{0.48\textwidth}{!}{
  \begin{tabular}{lcc}
  \toprule
  \textbf{Model} & \textbf{English-German} & \textbf{English-French}\\
    \midrule
    Transformer & 28.64 & 35.8\\
    Back-translation & 29.45 & 36.3\\
    \textsc{CsaNMT} & \bf 31.29 & \bf 38.6\\
  \bottomrule
\end{tabular}}
\end{center}
\setlength{\abovecaptionskip}{0.1cm}
\caption{BLEU scores [\%] on the IWSLT tasks. For fairer comparison, all the models are implemented by ourselves using the same backbone, and the extra monolingual corpora is not introduced into back-translation.}
\label{tbl:iwslt}
\end{table}

We further generalize the capability of the proposed \textsc{CsaNMT} to various low-resource machine translation tasks, including IWSLT14 English-German and IWSLT17 English-French. The details of the datasets and model configurations can be found in \textbf{Appendix~\ref{appendix:iwslt}}. Table~\ref{tbl:iwslt} shows the results of different models. Compared to the vanilla Transformer, the proposed \textsc{CsaNMT} improve the BLEU scores of the two tasks by 2.7 and 2.9 points, respectively. This result indicates that the claiming of the continuous semantic augmentation enriching the training corpora with very limited observed instances.

\section{Related Work}\label{sec:bg}

\noindent \textbf{Data Augmentation (DA)}~\cite{edunov-etal-2018-understanding,kobayashi-2018-contextual,gao-etal-2019-soft,khayrallah-etal-2020-simulated,pham2021meta} has been widely used in neural machine translation. The most popular one is the family of back-translation~\cite{sennrich-etal-2016-improving,NEURIPS2020_7221e5c8}, which utilizes a target-to-source model to translate monolingual target sentences back into the source language. Besides, constructing adversarial training instances with diverse literal forms via word replacing or embedding interpolating~\citep{wang-etal-2018-switchout,cheng-etal-2020-advaug} is beneficial to improve the generalization performance of NMT models.

\textbf{Vicinal Risk Minimization (VRM)}~\citep{Chapelle2020VRM} is another principle of data augmentation, in which DA is formalized as extracting additional pseudo samples from the vicinal distribution of observed instances. Typically the vicinity of each training instance is defined artificially according to the characteristics of the dataset (or task), such as color (scale, mixup) augmentation~\citep{simonyan2014very,Krizhevsky2012ImageNet,zhang2018mixup} in computer vision and adversarial augmentation with manifold neighborhoods~\citep{ng-etal-2020-ssmba,cheng2021self} in NLP. Our approach relates to VRM that involves with an adjacency semantic region as the vicinity manifold for each training instance.

\textbf{Sentence Representation Learning} is a well investigated area with dozens of methods~\citep{kiros2015skip,cer2018universal,yang-etal-2018-learning}. In recent years, the methods built on large pre-trained models~\citep{devlin-etal-2019-bert,conneau-etal-2020-unsupervised} have been widely used for learning sentence level representations~\citep{reimers-gurevych-2019-sentence,huang-etal-2019-unicoder,yang-etal-2019-sentence}. Our work is also related to the methods that aims at learning the universal representation~\cite{zhang-etal-2016-variational-neural,schwenk2017learning,yang-etal-2021-universal} for multiple semantically-equivalent sentences in NMT. In this context, \textbf{\textit{contrastive learning}} has become a popular paradigm in NLP~\citep{Kong2020A,DBLP:conf/iclr/ClarkLLM20,gao-etal-2021-simcse}. The most related work are \citet{wei2021on} and \citet{chi-etal-2021-infoxlm}, which suggested transforming cross-lingual sentences into a shared vector by contrastive objectives.


\section{Conclusion}
\vspace{-0.2cm}
We propose a novel data augmentation paradigm \textsc{CsaNMT}, which involves with an adjacency semantic region as the vicinity manifold for each training instance. This method is expected to make more unseen instances under generalization with very limited training data. The main components of \textsc{CsaNMT} consists of the tangential contrastive learning and the Mixed Gaussian Recurrent Chain (\textsc{Mgrc}) sampling. Experiments on both rich- and low-resource machine translation tasks demonstrate the effectiveness of our method.

In the future work, we would like to further study the vicinal risk minimization with the combination of multi-lingual aligned scenarios and large-scale monolingual data, and development it as a pure data augmentator merged into the vanilla Transformer.


\section*{Acknowledgments}
We would like to thank all of the anonymous reviewers (during ARR Oct. and ARR Dec.) for the
helpful comments. We also thank Baosong Yang and Dayiheng Liu for their instructive suggestions and invaluable help.

\bibliography{anthology,custom}
\bibliographystyle{acl_natbib}

\appendix
\section{Details of Rich-Resource Datasets}
\label{appendix:rich-dataset}

\begin{table*}[t]
  \begin{center}
  \resizebox{\textwidth}{!}{
  \begin{tabular}{lll}
  \toprule
  \textbf{Variants} & \textbf{Training Objective for the Semantic Encoder} & \textbf{Sampling Strategy for Obtaining Augmented Samples}\\
    \midrule
    1 & $\mathbb{E}_{(\mathbf{x}^{(i)},\mathbf{y}^{(i)}) \sim \mathcal{B}} \Bigg( \log \frac{e^{s \big( r_{x^{(i)}},r_{y^{(i)}} \big) }}{e^{s \big( r_{x^{(i)}},r_{y^{(i)}} \big) } + \sum_{j \& j \ne i}^{|\mathcal{B}|} \Big ( e^{s \big( r_{y^{(i)}},r_{y^{\prime(j)}} \big) } + e^{s \big( r_{x^{(i)}},r_{x^{\prime(j)}} \big) } \Big)} \Bigg)$ & $\omega^{(k)} \sim \eta \mathcal{N}\big(\mathbf{0},{\rm diag}(\mathcal{W}_{r}^2) \big) + (1.0-\eta) \mathcal{N}\big(\mathbf{0},\mathbf{1} \big)$\\
    2 & ditto & $\omega^{(k)} \sim \eta \mathcal{U}\big(-\mathcal{W}_{r},\mathcal{W}_{r} \big) + (1.0-\eta) \mathcal{U}\big(\bar{\mathbf{a}} -\mathbf{1},\mathbf{1} - \bar{\mathbf{a}} \big)$ \\
    & & where $\bar{\mathbf{a}} = \frac{1}{k-1}\sum_{i=1}^{k-1}\omega^{(i)}$\\
    3 & $\mathbb{E}_{(\mathbf{x}^{(i)},\mathbf{y}^{(i)}) \sim \mathcal{B}} \Big( -KL \big( p(r_{x^{(i)}}) \parallel q(r_{x^{(i)}},r_{y^{(i)}}) \big) \Big)$ & $\hat{r}_x = \mu + \epsilon \odot \sigma$\\
    & where $p(r_{x^{(i)}})\sim \mathcal{N}(\mu,\sigma^2)$ and $q(r_{x^{(i)}}, r_{y^{(i)}})\sim \mathcal{N}(\mu^{\prime},\sigma^{\prime 2})$ & where $\epsilon$ is a standard Gaussian noise\\
    4 & $\mathbb{E}_{(\mathbf{x}^{(i)},\mathbf{y}^{(i)}) \sim \mathcal{B}} \Big( \frac{r_{x^{(i)}}^T r_{y^{(i)}}}{\parallel r_{x^{(i)}} \parallel \cdot \parallel r_{y^{(i)}} \parallel} \Big)$ & $\omega^{(k)} \sim \eta \mathcal{N}\big(\mathbf{0},{\rm diag}(\mathcal{W}_{r}^2) \big) + (1.0-\eta) \mathcal{N}\Big(\frac{1}{k-1}\sum_{i=1}^{k-1}\omega^{(i)},\mathbf{1} \Big)$\\
  \bottomrule
\end{tabular}}
\end{center}
\caption{The variants of the training objective for the semantic encoder as well as the sampling strategy for obtaining augmented samples.}
\label{tbl:variants-imcmc-ctl}
\end{table*}

\begin{table*}[t]
  \begin{center}
  \resizebox{\textwidth}{!}{
  \begin{tabular}{lccccccc}
  \toprule
  \multirow{2}{*}{\textbf{Model}}& \multicolumn{4}{c }{\textbf{Noisy Inputs}}  & \multicolumn{3}{c}{\textbf{Translationese}} \\
  \cmidrule(r){2-5} \cmidrule(r){6-8}
  & \bf Original & \bf WS & \bf WD & \bf WR & ${\bf X \rightarrow Y^*}$ & ${\bf X^* \rightarrow Y}$ & ${\bf X^{**} \rightarrow Y^*}$\\
    \midrule
    Transformer (\textit{our implementation}) & 27.67 & 15.33 & 18.59 & 16.98 & 32.82 & 28.56 & 39.04 \\
    Back-Translation (\textit{our implementation}) & 29.25 & 17.20 & 20.44 & 18.71 & 33.07 & 29.73 & 39.86 \\
    \textsc{CsaNMT} & 30.16 & 20.14 & 23.76 & 21.66 & 34.62 & 30.70 & 41.64\\
  \bottomrule
\end{tabular}}
\end{center}
\caption{BLEU scores [\%] on noisy inputs and the translationese effect, in the WMT14 En$\rightarrow$De setup.}
\label{tbl:robustness}
\end{table*}

For the Zh$\rightarrow$En task, the LDC corpus \footnote{LDC2002E18, LDC2003E07, LDC2003E14, the Hansards portion of LDC2004T07-08 and LDC2005T06.} is taken into consideration, which consists of 1.25M sentence pairs with 27.9M Chinese words and 34.5M English words, respectively. The NIST 2006 dataset is used as the validation set for selecting the best model, and NIST 2002 (MT02), 2003 (MT03), 2004 (MT04), 2005 (MT05), 2008 (MT08) are used as the test sets. We created shared BPE (byte-pair-encoding~\citep{sennrich-etal-2016-neural}) codes with 60K merge operations to build two vocabularies comprising 47K Chinese sub-words and 30K English sub-words. For the En$\rightarrow$De task, we employ the popular WMT14 dataset, which consists of approximately 4.5M sentence pairs for training. We select \texttt{newstest2013} as the validation set and \texttt{newstest2014} as the test set. All sentences had been jointly byte-pair-encoded with 32K merge operations, which results in a shared source-target vocabulary of about 37K tokens. For the En$\rightarrow$Fr task, we use the significantly larger WMT14 dataset consisting of 36M sentence pairs. The combination of \{\texttt{newstest2012, 2013}\} was used for model selection and the experimental results were reported on \texttt{newstest2014}.

We use the Stanford segmenter~\citep{tseng-etal-2005-conditional} for Chinese word segmentation and apply the script \texttt{tokenizer.pl} of Moses~\citep{koehn-etal-2007-moses} for English, German and French tokenization. We measure the performance with the 4-gram BLEU score~\citep{papineni-etal-2002-bleu}. Both the case-sensitive tokenized BLEU (compued by \texttt{multi-bleu.pl}) and the detokenized sacrebleu\footnote{\url{https://github.com/mjpost/sacrebleu}} \citep{post-2018-call} are reported on the En$\rightarrow$De and En$\rightarrow$Fr tasks. The case-insensitive BLEU is reported on the Zh$\rightarrow$En task.

\section{Low-Resource Machine Translation}
\label{appendix:iwslt}


\paragraph{Datasets.} For IWSLT14 En$\rightarrow$De, there are $160k$ sentence pairs for training and $7584$ sentence pairs for validation. As in previous work~\cite{Ranzato2015Sequence,Zhu2020Incorporating}, the concatenation of dev2010, dev2012, test2010, test2011 and test2012 is used as the test set. For IWSLT17 En$\rightarrow$Fr, there are $236k$ sentence pairs for training and $10263$ for validation. The concatenation of test2010, test2011, test2012, test2013, test2014 and test2015 is used as the test set. We use a joint source and target vocabulary with $10k$ byte-pair-encoding (BPE) types~\cite{sennrich-etal-2016-neural} for above two tasks.

\paragraph{Model Settings.} The model configuration is \texttt{transformer\_iwslt}, representing a 6-layer model with embedding size 512 and FFN layer dimension 1024. We train all models using the Adam optimizer with adaptive learning rate schedule (warm-up step with 4K) as in~\cite{Vaswani2017Attention}. During inference, we use beam search with a beam size of $5$ and length penalty of $1.0$.

\section{Variants with Different Objectives or Sampling Strategies}
\label{appendix:objective-sampling}

Table~\ref{tbl:variants-imcmc-ctl} describes the details of four variants (introduced in Table~\ref{tbl:imcmc-ctl}, from row 2 to row 5) with different objectives or sampling strategies: (1) default tangential CTL in Eq. (\ref{eq:ctl}) + \textsc{Mgrc} w/o recurrent dependence, (2) default tangential CTL in Eq. (\ref{eq:ctl}) + \textsc{Mgrc} w uniform distribution, (3) variational inference~\citep{zhang-etal-2016-variational-neural} + Gaussian sampling, and (4) cosine similarity + default \textsc{Mgrc} sampling.

\section{Robustness on Noisy Inputs and Translationese}
\label{appendix:robustness}

In this section, we study the robustness of our \textsc{CsaNMT} towards both noisy inputs and the translationese effect~\citep{10.1093/llc/fqt031} on \textit{newstest2014} for the WMT14 English-German task.

\paragraph{Noisy Inputs.} Inspired by~\citep{gao-etal-2019-soft}, we construct noisy test sets via several strategies described as follows:
\begin{itemize}
    \item \textbf{Original}: the original testset without any manipulations;
    \item \textbf{WS}: word swap, randomly swap words in nearby positions within a window size 3~\citep{artetxe2018unsupervised,lample2018unsupervised};
    \item \textbf{WD}: word dropout, randomly drop words with a ratio of 15\%~\citep{iyyer-etal-2015-deep,lample2018unsupervised};
    \item \textbf{WR}: word replace, randomly replace word tokens with a placeholder token (e.g., \texttt{[UNK]})~\citep{xie2017data} or with a relevant (measured by the similarity of word embeddings) alternative~\citep{cheng-etal-2019-robust}. The replacement ratio also is 15\%.
\end{itemize}

\paragraph{Translationese Effect.} \citet{edunov-etal-2020-evaluation} pointed out that \textit{``back-translation (BT) suffers from the translationese effect. That is BT only shows significant improvements for test examples where the source itself is a translation, or translationese, while is ineffective to translate natural text''}. To test the effect of our method on translationese, we follow the same settings and testsets\footnote{\url{https://github.com/facebookresearch/evaluation-of-nmt-bt}} provided by~\citet{edunov-etal-2020-evaluation}:
\begin{itemize}
    \item natural source $\rightarrow$ translationese target (${\bf X \rightarrow Y^*}$);
    \item translationese source $\rightarrow$ natural target (${\bf X^* \rightarrow Y}$);
    \item round-trip translationese source $\rightarrow$ translationese target (${\bf X^{**} \rightarrow Y^*}$), where ${\bf X \rightarrow Y^* \rightarrow X^{**}}$.
\end{itemize}

\paragraph{Results.} As shown in Table~\ref{tbl:robustness}, our
approach shows better robustness over two baseline methods across various artificial noises. Moreover, \textsc{CsaNMT} consistently outperforms the baseline in all three translationese scenarios, the same is true for back-translation. However, \citet{edunov-etal-2020-evaluation} shows that BT improves only in the ${\bf X^* \rightarrow Y}$ scenario. Our explanation for the inconsistency is that BT without monolingual data in our setting benefits from the natural parallel data to deal with the translationese sources.

\onecolumn

\section{Codes of tangential contrastive learning and \textsc{Mgrc} sampling}
\label{appendix:codes}

\subsection{Tangential Contrastive Learning}
\begin{lstlisting}[language=Python]
# src_embedding: [batch_size, 1, hidden_size]
# trg_embedding: [batch_size, 1, hidden_size]
def get_ctl_loss(src_embedding, trg_embedding, dynamic_coefficient):
    batch_size = tf.shape(src_embedding)[0]
    def get_ctl_logits(query, keys):
        # expand_query: [batch_size, batch_size, hidden_size]
        # expand_keys: [batch_size, batch_size, hidden_size]
        # the current ref is the positive key, while others in the training batch are negative ones
        expand_query = tf.tile(query, [1, batch_size, 1])
        expand_keys = tf.tile(tf.transpose(keys, [1,0,2]), [batch_size, 1, 1])

        # distances between queries and positive keys
        d_pos = tf.sqrt(tf.reduce_sum(tf.pow(query - keys, 2.0), axis=-1)) # [batch_size, 1]
        d_pos = tf.tile(d_pos, [1, batch_size]) # [batch_size, batch_size]
        d_neg = tf.sqrt(tf.reduce_sum(tf.pow(expand_query - expand_keys, 2.0), axis=-1)) # [batch_size, batch_size]

        lambda_coefficient = (d_pos / d_neg) ** dynamic_coefficient
        hardness_masks = tf.cast(tf.greater(d_neg, d_pos), dtype=tf.float32)

        hard_keys =(expand_query + tf.expand_dims(lambda_coefficient, axis=2) * (expand_keys - expand_query)) * \
          tf.expand_dims(hardness_masks, axis=2) + expand_keys * tf.expand_dims(1.0 - hardness_masks, axis=2)

        logits = tf.matmul(query, hard_keys, transpose_b=True) # [batch_size, 1, batch_size]
        return logits

    logits_src_trg = get_ctl_logits(src_embedding, trg_embedding)
    logits_trg_src = get_ctl_logits(trg_embedding, src_embedding) + tf.expand_dims(tf.matrix_band_part(tf.ones([batch_size, batch_size]), 0, 0) * -1e9, axis=1)
    logits = tf.concat([logits_src_trg, logits_trg_src], axis=2) # [batch_size, 1, 2*batch_size]

    labels = tf.expand_dims(tf.range(batch_size, dtype=tf.int32), axis=1)
    labels = tf.one_hot(labels, depth=2*batch_size, on_value=1.0, off_value=0.0)

    cross_entropy_fn = tf.nn.softmax_cross_entropy_with_logits
    
    loss = tf.reduce_mean(cross_entropy_fn(logits=logits, labels=labels))
    return loss
\end{lstlisting}

\subsection{\textsc{Mgrc} Sampling}
\begin{lstlisting}[language=Python]
# src_embedding: [batch_size, hidden_size]
# trg_embedding: [batch_size, hidden_size]
# default: K=20 and eta = 0.6 
def mgrc_sampling(src_embedding, trg_embedding, K, eta):
    batch_size = tf.shape(src_embedding)[0]
    def get_samples(x_vector, y_vector):
        bias_vector = y_vector - x_vector
        W_r = (tf.abs(bias_vector) - tf.reduce_min(tf.abs(bias_vector), axis=1, keep_dims=True)) / \
              (tf.reduce_max(tf.abs(bias_vector), 1, keep_dims=True) - tf.reduce_min(tf.abs(bias_vector), 1, keep_dims=True))

        # initializing the set of samples
        R = []
        omega = tf.random_normal(tf.shape(bias_vector), 0, W_r)
        sample = x_vector + tf.multiply(omega, bias_vector)
        R.append(sample)

        for i in xrange(1, K):
            chain = [tf.expand_dims(item, axis=1) for item in R[:i]]
            average_omega = tf.reduce_mean(tf.concat(chain, axis=1), axis=1)
            omega = eta * tf.random_normal(tf.shape(bias_vector), 0, W_r) + \
                    (1.0 - eta) * tf.random_normal(tf.shape(bias_vector), average_omega, 1.0)
            sample = x_vector + tf.multiply(omega, bias_vector)
            R.append(sample)

        return R

    x_sample = get_samples(src_embedding, trg_embedding)
    y_sample = get_samples(trg_embedding, src_embedding)
    return x_sample.extend(y_sample)
\end{lstlisting}

\end{document}